\documentclass[11pt]{article}
\pdfoutput=1
\usepackage[utf8]{inputenc}  
\usepackage[T1]{fontenc}  
\usepackage{CJK}  

\usepackage{newtx}
\usepackage{amsmath} 
\usepackage[normalem]{ulem} 
\usepackage[colorlinks=true, linkcolor=blue, urlcolor=blue, citecolor=blue]{hyperref}

\let\oldHref\href
\renewcommand{\href}[2]{\oldHref{#1}{\uline{#2}}}
\usepackage[final]{acl}
\usepackage{times}
\usepackage{latexsym}
\usepackage[T1]{fontenc}

\usepackage{microtype}

\usepackage{inconsolata}

\usepackage{todonotes}

\title{Modeling Bilingual Sentence Processing: Evaluating RNN and Transformer Architectures for Cross-Language Structural Priming} 

\author{First Author \\
  Affiliation / Address line 1 \\
  Affiliation / Address line 2 \\
  Affiliation / Address line 3 \\
  \texttt{email@domain} \\\And
  Second Author \\
  Affiliation / Address line 1 \\
  Affiliation / Address line 2 \\
  Affiliation / Address line 3 \\
  \texttt{email@domain} \\}

\author{Demi Zhang, \ Bushi Xiao, \ Chao Gao, Sangpil Youm, Bonnie Dorr\\
  University of Florida\\ \texttt{\{zhang.yidan, xiaobushi, gao.chao, youms, bonniejdorr\}@ufl.edu}\\}
  
\begin{document}

\maketitle

\begin{abstract}

This study evaluates the performance of Recurrent Neural Network (RNN) and Transformer models in replicating cross-language structural priming, a key indicator of abstract grammatical representations in human language processing. Focusing on Chinese-English priming, which involves two typologically distinct languages, we examine how these models handle the robust phenomenon of structural priming, where exposure to a particular sentence structure increases the likelihood of selecting a similar structure subsequently. Our findings indicate that transformers outperform RNNs in generating primed sentence structures, with accuracy rates that exceed 25.84\% to 33. 33\%. This challenges the conventional belief that human sentence processing primarily involves recurrent and immediate processing and suggests a role for cue-based retrieval mechanisms. This work contributes to our understanding of how computational models may reflect human cognitive processes across diverse language families.

\end{abstract}



\section{Introduction}

Structural priming refers to the phenomenon where encountering a specific syntactic structure boosts the probability of generating or understanding sentences with a comparable structure \citep{pickering2008structural}. It serves as a valuable method for exploring the capabilities of language models and probing their internal states and their potential relation to human sentence processing.

Studies show that Recurrent Neural Networks (RNN), particularly Gated Recurrent Unit models (GRU), have been pivotal in modeling human sentence processing, including structural priming \citep{frank2019neural}. Meanwhile, transformers also demonstrate structural priming ability similar to that of humans \citep{sinclair-etal-2022-structural}. This suggests the representations learned by the models may capture not only sequential structure but also some degree of hierarchical syntactic information.

That said, to our knowledge, no study has 
compared these models'
ability to syntactically prime across two typologically distant languages. In the current study, we address this gap by comparing the models' ability to prime syntactically across two languages from vastly different families.

\begin{figure}
\centering
\scriptsize
\includegraphics[width=0.48\textwidth]{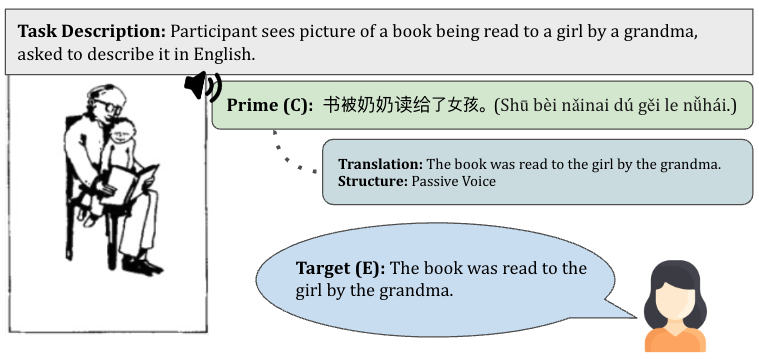}
\vspace*{-.1in}
\caption{Cross-language structure priming of human participant: \textit{C} denotes Chinese, \textit{E} denotes English.
}
\label{fig:figure_example}
\vspace*{-.2in}
\end{figure}

Consider a case where a human participant reads 
a passive Chinese (C) sentence 
and is then asked to describe a separate picture in English (E) (see Figure~\ref{fig:figure_example}). 
Here, the passive sentence C 
influences the structure of the target sentence E, leading the participant to use passive voice in their description. 


Our study explores structural priming in translation models, highlighting 
their 
ability to generate syntactically diverse English outputs from Chinese inputs. A key contribution is a set of insights into syntactic representation across typologically distinct languages in computation models. We demonstrate that transformers outperform RNNs in generating primed sentence structures, challenging the belief that human sentence processing relies mainly on recurrent and immediate processing. 

The next section reviews work on
cross-linguistic priming. Section~\ref{sec:current-study} introduces our study, 
exploring insights into syntactic representation across typologically distinct languages in computational models.
Section~\ref{sec:data} 
introduces a newly designed test set 
to evaluate our models. 
Section~\ref{sec:language-models} details
the implementation and training of two distinct models. Section~\ref{sec:experimental_setup} discusses
the design of our experimental setup, followed
by a comprehensive analysis and interpretation of our results.

\section{Related Work}
\label{sec:related}
This section focuses on work related to cross-linguistic priming, as exemplified in Figure~\ref{fig:figure_example}. Prior experiments induce
cross-linguistic structural priming 
by instructing bilingual participants to use two languages: presenting primes in one language and eliciting targets in another. These studies show that specific sentence structures in one language can influence the use of similar structures in the other language \citep{hartsuiker_is_2004}. 

Computational modeling studies have 
shown that RNNs exhibit structural priming effects akin to those observed in human bilinguals \citep{frank2021cross}. These 
models process sequential information through recurrence, a 
feature thought to resemble human cognitive processing. The 
emergence
of such priming effects in language models suggests that they develop implicit syntactic representations that
resemble
those employed by 
human language systems \citep{linzen2021syntactic}.

However, the transformer model, which uses self-attention mechanisms instead of recurrence, challenges this notion. The transformer's ability to directly access past input information, regardless of temporal distance, offers a fundamentally different approach from RNNs. The effectiveness of transformers in various NLP tasks makes us wonder if they can emulate RNNs in modeling cross-language structural priming.

The current study is inspired by two prior
studies. \citet{merkx_human_2021} 
compare 
transformer and RNN models' ability to
account
for measures of monolingual (English) human reading effort. 
They show that transformers outperform RNNs in explaining self-paced reading times and neural activity during 
English sentence reading, challenging the widely held idea that human sentence processing relies on recurrent and immediate processing. 
Their study is monolingual and English-centric. 
\citet{frank2021cross} investigates cross-language structural priming, finding
that RNNs
trained on English-Dutch
sentences account for garden-path effects and are sensitive to structural priming, within and between languages.

Recent studies on structural priming in neural language models have shown significant progress, with researchers quantifying this phenomenon using various methods across different languages. \citet{prasad-etal-2019-using} 
demonstrate that LSTM language models can hierarchically organize syntactic representations in a manner that reflects abstract sentence properties. \citet{sinclair-etal-2022-structural} show
that Transformer models exhibit structural priming, suggesting these models capture both sequential and hierarchical syntactic information. 

\citet{michaelov_structural_2023} 
provide evidence that large multilingual language models possess abstract grammatical representations that influence text generation similarly across different languages. Together, these findings underscore the capacity of neural models to develop and apply structural abstractions, contributing to a deeper understanding of language processing in AI.

\section{The Current Study}
\label{sec:current-study}
Our study examines structural priming in translation models, demonstrating
their capability to
generate syntactically diverse English outputs from Chinese inputs. This approach 
offers insights into syntactic representations across typologically distinct languages in computational models. 

To compare
RNNs and transformers 
in
their ability to
model cross-language structural priming, 
we 
adopt a new approach.
While \citet{frank2021cross} trains models on comprehension, where a longer response time indicates greater difficulty in 
understanding 
a new sentence
and thus a weaker priming effect,
the current study 
focuses on production. Here, the structure of 
each generated sentence
is compared with 
that 
of the input sentence to 
assess the presence of a priming effect.

\begin{figure}
    \centering
    \includegraphics[width=7.3cm,height=4.7cm]
    {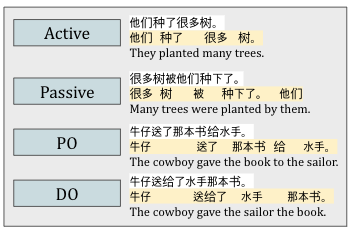}
       \vspace*{-.1in}
    \caption{Example of Active, Passive, Propositional Object (PO), and Double Object (DO). White highlighted sentence is original Chinese sentence, and yellow highlighted Sentence is word-to-word mapping between Chinese and English. }
    \label{fig:ex_sentence}
    \vspace*{-.2in}
\end{figure}

\begin{figure}
    \centering
    \includegraphics[width=7.3cm,height=8.3cm]
    {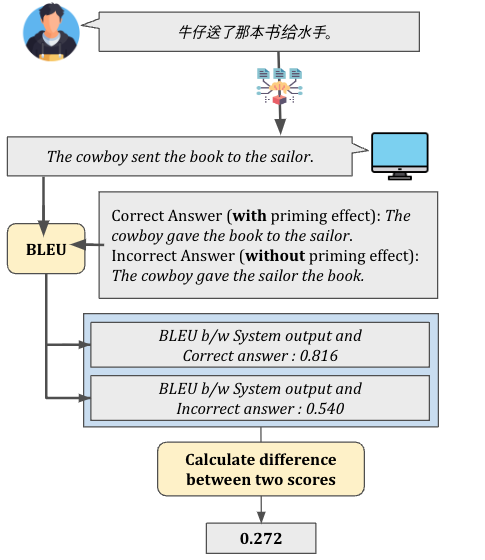}
    \caption{Example of test phase and evaluation process.}
    \label{fig:ex_test}
    \vspace*{-.2in}
\end{figure}

As shown in Figure~\ref{fig:ex_sentence}, Chinese has
equivalents 
for structures that are
passive (e.g., \textit{Many trees were planted by them}) and active (e.g., \textit{They planted many trees}). It also includes structures
for prepositional objects (e.g., \textit{The cowboy gave the book to the sailor}) and double objects 
(e.g., \textit{The cowboy gave the sailor the book}). In 
our study, the input sentence is in Chinese and system output is
an English version of the sentence. BLEU scores are calculated between the system-generated 
English sentence and both a ``correct'' 
English sentence that shares the structure with 
the Chinese input%
and an ``incorrect'' 
sentence. We then calculate the difference between the two BLEU scores, as depicted in Figure ~\ref{fig:ex_test}.

Another novel aspect of our study is 
the selection of two languages from
vastly 
divergent language families, 
challenging the models to develop abstract representations for distinct structures.

\begin{figure*}[t]
  \begin{center}  
  \includegraphics[width=0.85\linewidth]{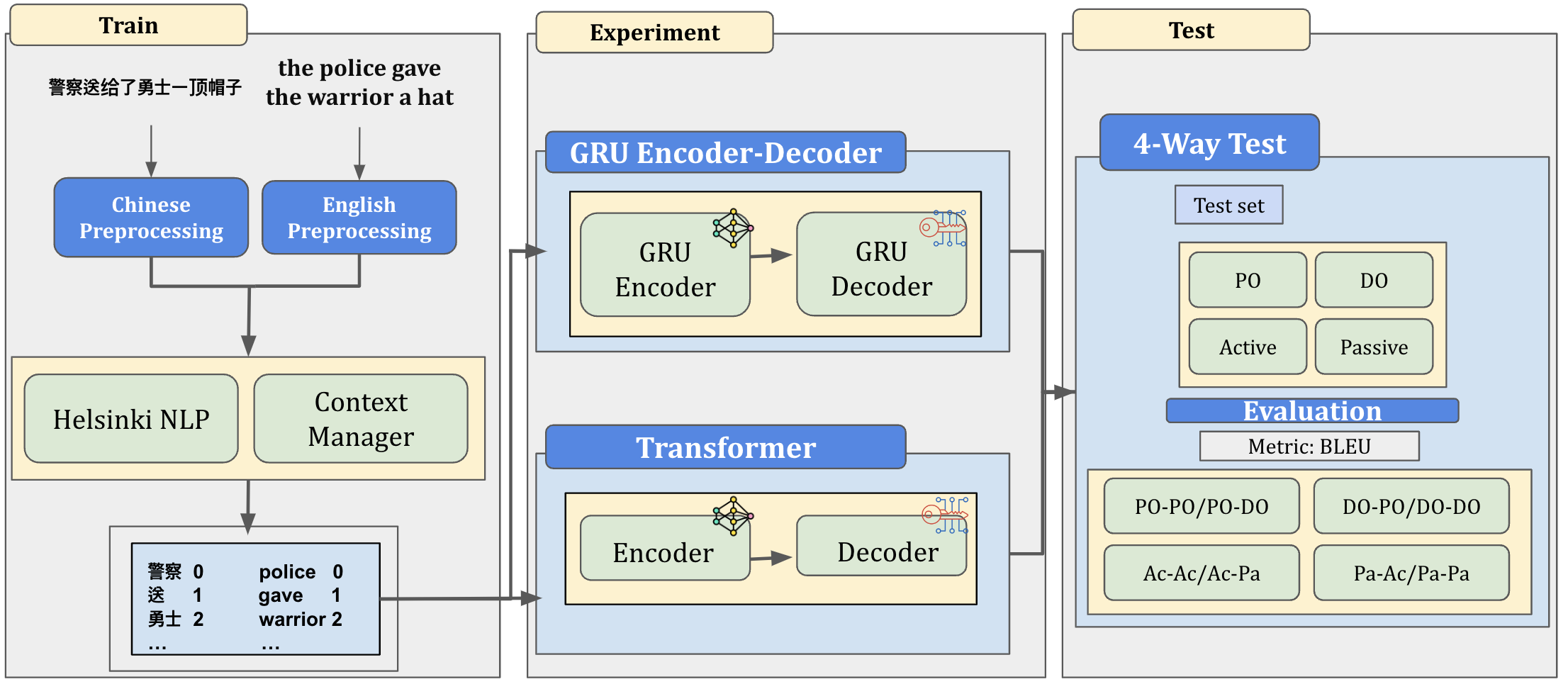}\\
  \vspace*{-.1in}
 \caption{The 
 workflow of the study 
 includes
 PO (Propositional Object), DO 
 (Double Object), 
 Ac 
 (Active), and Pa 
 (Passive).
 In
 the training phase, 
 raw bilingual data are preprocessed 
 to generate token pairs. In the experiment phase, we employ transformer and RNN-based encoder-decoder architectures. In the testing phase, we evaluate the model's performance 
 across four sentence structures 
 using the BLEU metric.}
 \label{architecture}
   \vspace*{-.2in}
 \end{center}
\end{figure*}


\section{Data Preparation}
\label{sec:data}

We select and process a Chinese-English  corpus which contains 
5.2 million Chinese-English
parallel sentence pairs \citep{bright_xu_2019_3402023}.\footnote{The source can be found at \url{https://drive.google.com/file/d/1EX8eE5YWBxCaohBO8Fh4e2j3b9C2bTVQ/view?pli=1} }

We employ a DataLoader \footnote{Our Dataloader is supported by PyTorch, referencing its license located at \url{https://github.com/pytorch/pytorch/blob/main/LICENSE}} to facilitate batch processing, transforming text into token IDs suitable for model interpretation. We then use the Helsinki-NLP tokenizer \cite{TiedemannThottingal:EAMT2020}\footnote{Helsinki-NLP is licensed under the MIT license. For more details, see here: \url{https://github.com/Helsinki-NLP/Opus-MT/blob/master/LICENSE}} to map
Chinese to English,
accommodating over a thousand models for diverse language pairs. 

The tokenizer, by default, processes text 
based on source language settings. To correctly encode target language text, the context manager must be set to use the
target tokenizer. Without this, the source language tokenizer would be incorrectly applied 
to the target text, leading to poor tokenization results, such as improper
word splitting for words not recognized
in the source language.

In sequence-to-sequence models, assigning a value of -100 to padding tokens ensures they are excluded from loss calculations.
This setup is crucial for effective model training, 
enabling 
precise adjustment of model parameters based on the tokenized input and target sequences.
Proper
data formatting
through this preprocessing step 
facilitates optimal training outcomes.

We also design a test dataset, initially sampling 
five sentences for each of the four 
sentence structures (Active Voice, Passive Voice, Prepositional Object, and Double Object) 
from the Cross-language Structural Priming Corpus \cite{michaelov_structural_2023}. To augment the data, 
we employ a LLM, ChatGPT 3.5 \cite{openai_gpt}, 
By providing a one-shot learning
prompt, we expand each set to 30 sentences, resulting in a total of 120 sentences for our test dataset:
\begin{center}
\vspace*{-.15in}
\begin{tabular}{p{3in}}
Generate 30 sentences with the following structure: \textit{The cowboy gave the book to the sailor.} Replace all the words while keeping the sentence structure the same.
\end{tabular}
\end{center}
In our test set, each Chinese sentence is paired with a correct and an incorrect English sentence. 

Subsequently, a bilingual annotator proficient in both Mandarin and English carefully reviews
the sentence outputs generated by the LLM, ensuring that each triplet comprises
translation equivalents. The review also confirms
that only the `correct' answer maintains
syntactic alignment with the original Chinese sentence.
\section{Language Models}
\label{sec:language-models}

We implement both a transformer model and an RNN model to handle sequence-to-sequence tasks using the encoder-decoder architecture. (See Experiment of Figure~\ref{architecture}.) This architecture supports the processing of
both input sequences 
and output sequences of varying lengths, which is crucial for 
accommodating 
sentences with different structures yet similar meanings. 
This section
explores why these language models can assist us 
identify structural priming.
We train and test our RNN model and transformer using AMD EPYC 75F3 8-Core Processor and 1 NVIDIA A100 GPU.

\subsection{Multi-head Attention in Transformer}

In the transformer model, we 
use the self-attention mechanism (AttModel) to 
capture sentence structure.
This mechanism
identifies dependencies between different positions and adjusts
the representation of each word based on its relationship with others, thus facilitating the learning of sentence structure. 
Following \citet{vaswani2017attention}, \newline
\begin{equation} \label{eq:pos1}
\text{Attention}(Q, K, V) = \text{softmax}\left(\frac{QK^T}{\sqrt{d_k}}\right)V
\end{equation}
where \(Q, K, V\) are obtained through linear transformations of an input sequence of text, each with its own learnable weight matrix. In the encoder part of model, \(Q, K, V\) comes from the same source sequence, while in the decoder,
\(Q\) comes from the target sequence, and \(K\) and \(V\) come from the encoder's output.
Since the computation of \(Q\), \(K\), and \(V\) requires processing the entire input sentence, the model can simultaneously focus on all positions and capture the sentence's structure. 

In the decoder part of the transformer model, 
multiple attention heads 
capture 
different levels of sentence features, leading to a more comprehensive representation of sentence structure. Each attention head specializes in capturing specific semantic relationships, such as word dependencies and distance relationships. 

This approach enhances the model's ability to comprehend the intricacies of sentence structure. The equation is as follows:
\begin{equation} \label{eq:pos2}
\text{MH}(Q, K, V) = \text{Concat}(head_1, \ldots, head_h) \cdot W^O
\end{equation}
where \(W^O\) is the weight matrix 
to be trained, and \(head_1, \ldots, head_h\), computed through 
equation \ref{eq:pos1},
represent
the attention weights of each head (we 
use 8 heads). \text{Concat} is the operation of 
joining tensors along their last dimension.

We also 
prioritize the choice of
positional encoding method. While the common method involves using sine and cosine functions, 
we opt for learnable positional embeddings.
We believe this approach offers more advantages for learning structural priming,
as it
helps our model better understand and encode the relative positions of words within a sentence. 

In contrast to the fixed positional encoding, learnable positional embeddings assign different weights to different positions, emphasizing the relevant positional information that contributes to the priming effect. This enables the model to capture more intricate positional relationships and dependencies specific to the task of structural priming.

\subsection{GRU Encoder and GRU Decoder}

Some studies \cite{zhou2018rnn} show 
that RNNs can preserve sentence structure and 
facilitate identification of structural priming environments.
Their sequential nature allows them to process input tokens based on 
the context of the entire sentence.
As each token is processed, the 
RNN's hidden state is updated, retaining information about preceding tokens and their contextual relevance. This sequential processing enables the model to capture word dependency relationships, thereby preserving the structural integrity of the sentence. 
Summarizing:
\vspace*{-.1in}
\begin{equation} \label{eq:pos3}
\text{State}(dh_i, c_i), p = f(\text{State}(dh_{i-1}, c_{i-1}),  m)
\vspace*{-.05in}
\end{equation}

The function \(f\) refers to the hidden layer of the RNN model, which is a neural network. It takes the previous layer's 
\text{State} {i-1} and the output vector from the previous time step \(m\) as input, and outputs the next layer's 
\text{State} {i} and prediction value \(p\) until it encounters the termination symbol. 
Here, \(dh\) signifies the hidden state of the RNN unit in decoder, tasked with capturing pertinent information
from the input sequence. In the initial decoder step, \(dh\) embodies the 
final output state of the encoder.
In subsequent decoder steps, \(dh\) denotes 
the preceding RNN unit's output. 

To address the challenge of
not being able to retain
the entire sentence structure, we introduce the attention mechanism. 
This feature
of the RNN model 
enables it
to focus more on
the parts of the input sequence that are
most relevant to the current output, thereby enhancing prediction accuracy.
Its 
potential for predicting structural patterns stems from the attention mechanism's
ability to capture dependencies 
within sequential
data and 
to 
leverage these for better predictions. As shown in 
equation \ref{eq:pos3}, 
\(c\) denotes
the attention, and its calculation
is as follows:
\vspace*{-.1in}
\begin{equation} \label{eq:pos4}
\alpha_{i} = g(eh_i, dh_0)
\end{equation}

As 
before, \(dh_0\) denotes the final state of
the encoder and \(eh\) signifies the hidden state of the each RNN unit in the encoder. Function \(g\) is used to calculate the weight \(\alpha_{i}\) of \(eh_i\) in the final state \(dh_0\). As a result, 
the attention \(c\) is obtained by combining all 
previous states:
\vspace*{-.05in}
\begin{equation} \label{eq:pos5}
c_{i} = \sum(\alpha_{i} * dh_{i})
\end{equation}
calculated by summing the products of the weight \(\alpha\) and the 
decoder state \(dh\).

Our study utilizes
a variant of RNNs known as
the Gated Recurrent Unit (GRU). The GRU encoder and 
decoder are
gating mechanisms that
effectively manage
long-distance dependencies and 
mitigate
the vanishing gradient problem. Additionally, 
GRUs possess
fewer parameters and demonstrate higher computational efficiency.

Following 
\citet{dey2017gate}, we define the gate mechanism in two parts:
\begin{itemize}
\vspace*{-.05in}
\item 
Update Gate:
$z_t = \sigma(W_z x_t + U_z h_{t-1} + b_z)$
\end{itemize}
The update gate \( z_t \) in the encoder controls the blending of the current input \( x_t \) and the previous hidden state \( h_{t-1} \).  
In the decoder, the
update gate
regulates the interaction between the current input and the previous decoder state, allowing
the model to selectively incorporate relevant information from the input when generating the output.
\begin{itemize}
\vspace*{-.05in}
\item Reset Gate:
$r_t = \sigma(W_r x_t + U_r h_{t-1} + b_r)$
\end{itemize}

The reset gate \( r_t \) in the encoder regulates the interaction between
the current input \( x_t \) 
and
the previous hidden state \( h_{t-1} \).
In the decoder, the
reset gate
governs how the current input interacts with the
previous decoder state. 
This allows the model to selectively forget certain parts of the input information captured by the encoder. This helps
the decoder to generate outputs that are less influenced by outdated information from the input sequence.


\section{Experimental Setup}
\label{sec:experimental_setup}
Since structural priming effects are sometimes not symmetrical, our study only includes a structural priming experiment with Mandarin to English bilinguals, while existing literature strongly supports the presence of structural priming effects in both language directions. 

To assess the 
effectiveness
of our model in Chinese-English, we adopt the standard bilingual evaluation understudy (BLEU) metric \citep{papineni2002bleu}, which 
ranges from 0 to 1, indicating the similarity of predicted text against target text: 
$$\text{BLEU} = \text{BP} \cdot \exp\left(\sum_{n=1}^{N} w_n \log p_n\right)$$
Here,
$N$ is the maximum n-gram order (typically 4),
$w_n$ is the weight assigned to each n-gram precision score (with $\sum_{n=1}^{N} w_n = 1$), $p_n$ is the precision score for n-grams of order $n$, and
$\text{BP}$ is the brevity penalty which penalizes shorter results.

After generating 
predicted outcomes and 
assembling 
a test set, we analyze the relationship between 
predictions and four 
types of reference sentences: (1)~correct mappings with the same structure; (2)~semantically similar but structurally different sentences; (3)~semantically different but structurally identical sentences; and (4)~sentences 
that differ
both semantically and structurally.

We divide 
the comparisons into two 
groups based on semantic similarity. In the first group 
of sentences with identical meanings, we hypothesize that effective structural priming would result in higher BLEU scores
between the predicted sentences and the reference sentences that share the
same structure, compared to
those with different structures. This comparison aims to establish whether the model 
prefers to reproduce structures that are syntactically aligned with the ground truths when the semantic content remains
constant.

The second category, 
with sentences differing
in meaning,
is 
crucial for demonstrating structural priming, as it eliminates 
the influence of semantic similarity.
If sentences with identical structures receive 
higher BLEU scores than
those with different structures, 
it suggests
the model's predictions are 
driven by structure,
regardless
of semantic changes.

This methodology rigorously tests for
structural priming, 
offering insights into how 
models process and replicate
language structures.


\section{Results and Analyses}
\label{sec:results}
We present the performance of the GRU-based RNN and standard transformer model \citep{vaswani2017attention} 
demonstrating their crosslingual structural priming effect in Chinese-English 
scenarios. 


\subsection{Structural Priming Performance}

\begin{figure}
\centering
\scriptsize
\includegraphics[width=0.5\textwidth]{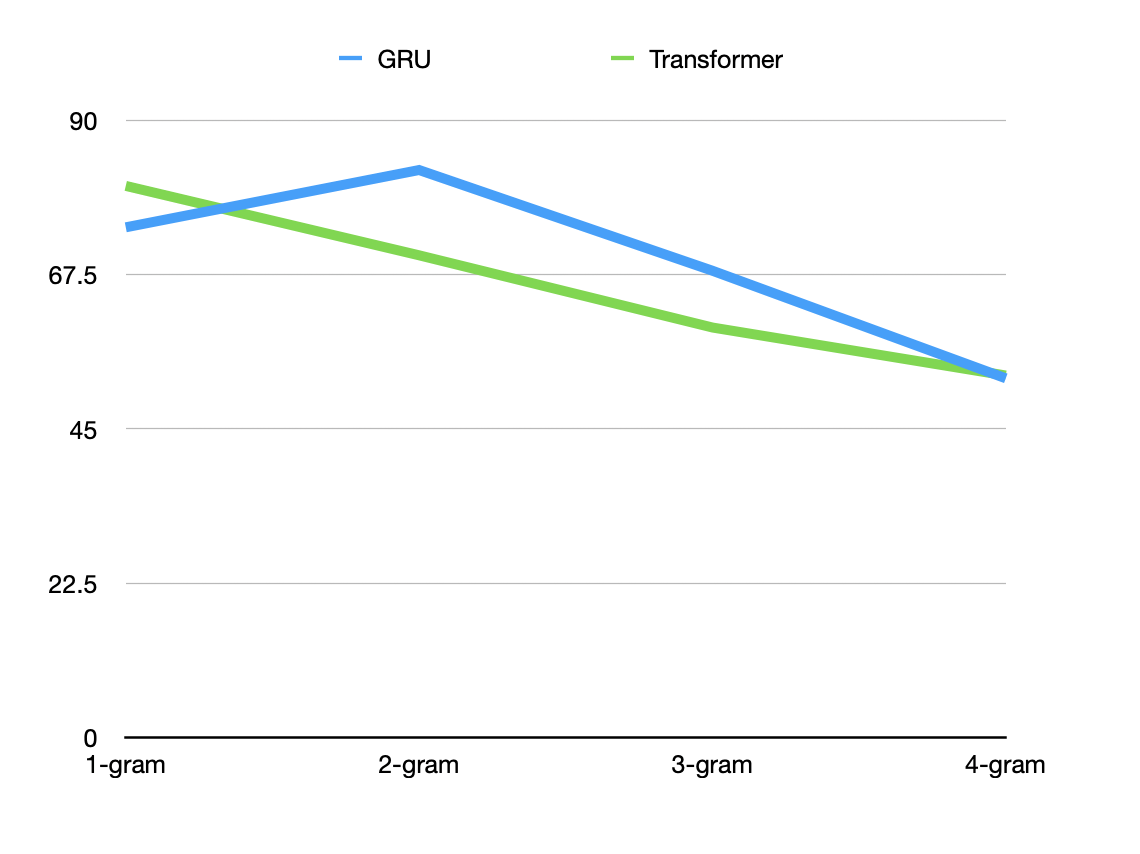}
\vspace*{-.2in}
\caption{BLEU Score for standard structural priming. Comparison of ground truth datasets for testing and calibration.
}
\label{BLEU_structure}
\vspace*{-.15in}
\end{figure}

Our 
analysis reveals
that, although
both models achieve competitive BLEU scores, the transformer model 
shows a slight edge
in handling complex sentence structures.
Figure~\ref{BLEU_structure}
shows that, when the training dataset is sufficiently large, both models attain high
predicted BLEU scores for sentence segments. 
Figures~\ref{BLEU_structure}--\ref{BLEU_wrong}
use BLEU scores,
common in translation and relevant to structural priming, where identical structures yield higher scores \citep{Lopez2008}.

\subsection{Crosslingual Structural Priming Effect}

\begin{figure}[t]
\centering
\scriptsize
\includegraphics[width=0.5\textwidth]{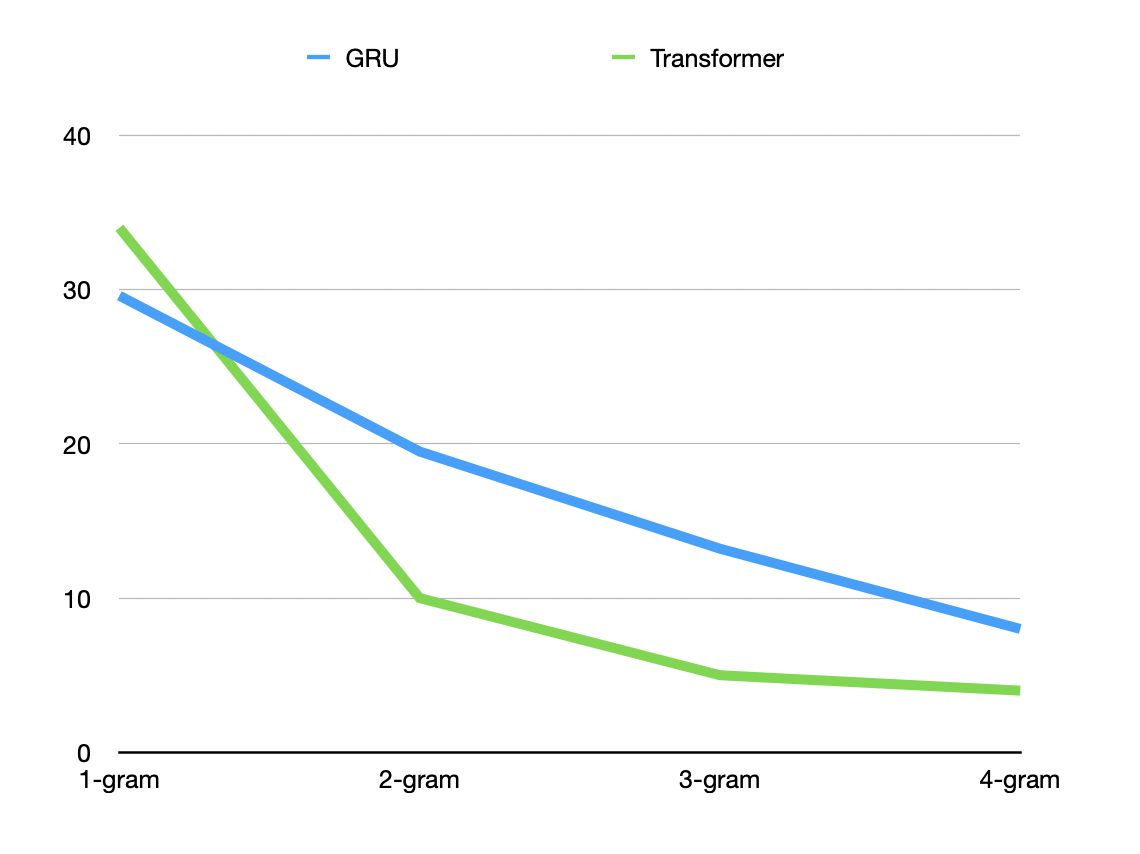}
\vspace*{-.2in}
\caption{BLEU Score for wrong priming. Comparison between predictions for cross-language priming via average BLEU Score.}
\label{BLEU_correct}
\centering
\scriptsize
\includegraphics[width=0.5\textwidth]{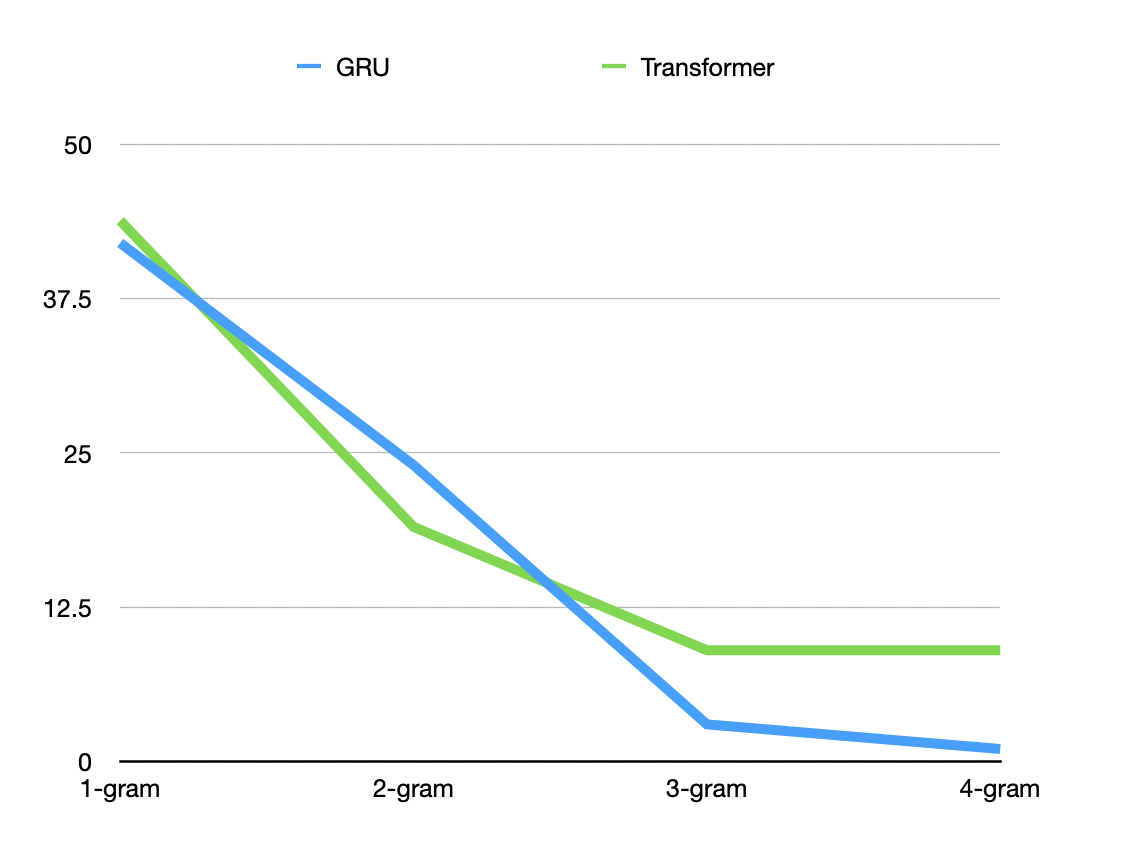}
\vspace*{-.2in}
\caption{BLEU Score for correct priming. Comparison between predictions for 
opposite cross-language priming via average BLEU Score.
}
\label{BLEU_wrong}
\vspace*{-.15in}
\end{figure}


Our 
crosslingual structural priming exploration reveals
a noteworthy pattern: both models facilitate the use of target-language syntactic structures influenced by the source language. However, the transformer model displays a 
stronger
priming effect, 
suggesting
a potential edge in mimicking human-like syntactic adaptation in bilingual contexts.

Figure~\ref{BLEU_correct} and Figure~\ref{BLEU_wrong} show 
BLEU scores for machine-generated predictions with correct or opposite priming test sets. This representation allows for a more direct comparison with the results from machine translation models, facilitating
a broader discussion 
regarding language structure in neural networks. From these we gain insights into model performance by
evaluating how closely 
predictions align with the correct structures
(e.g., Active-Active, DO-DO) versus opposite
structures (e.g., Active-Passive, PO-DO). Higher BLEU scores against the correct priming 
sets indicate better structural alignment,
whereas higher scores against 
opposite priming 
sets suggest deviations.
For 1-gram and 2-gram comparisons, GRU and transformer models perform similarly. However,
as n-grams increase, the transformer shows higher BLEU scores, indicating a closer alignment with incorrect structures. 
Overall, GRU outperforms the transformer in avoiding opposite priming (see Figure~\ref{BLEU_wrong}).
 
These results 
show that, when evaluated against the correct priming test sets, the transformer model 
performs similarly 
to GRU (see Figure~\ref{BLEU_correct}), with slight improvements 
as the n-gram size increases. However,
GRU generally outperforms the transformer 
compared to opposite priming (see Figure~\ref{BLEU_wrong}). Given that this 
involves
``incorrect'' priming, GRU aligns more closely with the opposite priming test set. Since the transformer shows a larger gap between correct and incorrect BLEU scores,
We infer that 
it
adheres more closely to the appropriate structural priming.

In a previous study, \citet{michaelov_structural_2023} examine the presence of structural priming by comparing the proportion of target sentences produced after different types of priming statements. Similarly, in our study, 
we prime the language model with 
a specific sentence for each experimental item and then calculate the normalized probabilities 
for the two target sentences. 
These normalized probabilities are computed as follows:

First, calculate the raw probability of each target sentence given the priming sentence:
\vspace*{-.2in}
\begin{center}
\small
\begin{align*}
P(\text{DO Target} | \text{DO Prime}) \\
P(\text{PO Target} | \text{PO Prime}) \\
P(\text{DO Target} | \text{PO Prime}) \\
P(\text{PO Target} | \text{DO Prime})
\end{align*}
\vspace*{-.2in}
\end{center}
And the same method for:
\vspace*{-.2in}
\begin{center}
\small
\begin{align*}
P(\text{Active Target} | \text{Active Prime}) \\
P(\text{Passive Target} | \text{Passive Prime}) \\
P(\text{Active Target} | \text{Passive Prime}) \\
P(\text{Passive Target} | \text{Active Prime})
\end{align*}
\vspace*{-.2in}
\end{center}

These probabilities are then normalized to calculate the conditional probability of the target sentence, assuming
the model outputs 
one of the two target sentences. Taking DO | PO as example:
\begin{center}
\vspace*{-.1in}
\scriptsize
\begin{multline*}
P_N(\text{Target} | \text{Prime}) =
\frac{P(\text{Target} | \text{Prime})}{P(\text{DO Target} | \text{Prime}) + P(\text{PO Target} | \text{Prime})}
\end{multline*}
\end{center}

\begin{figure}
\centering
\scriptsize
\includegraphics[width=0.47\textwidth]{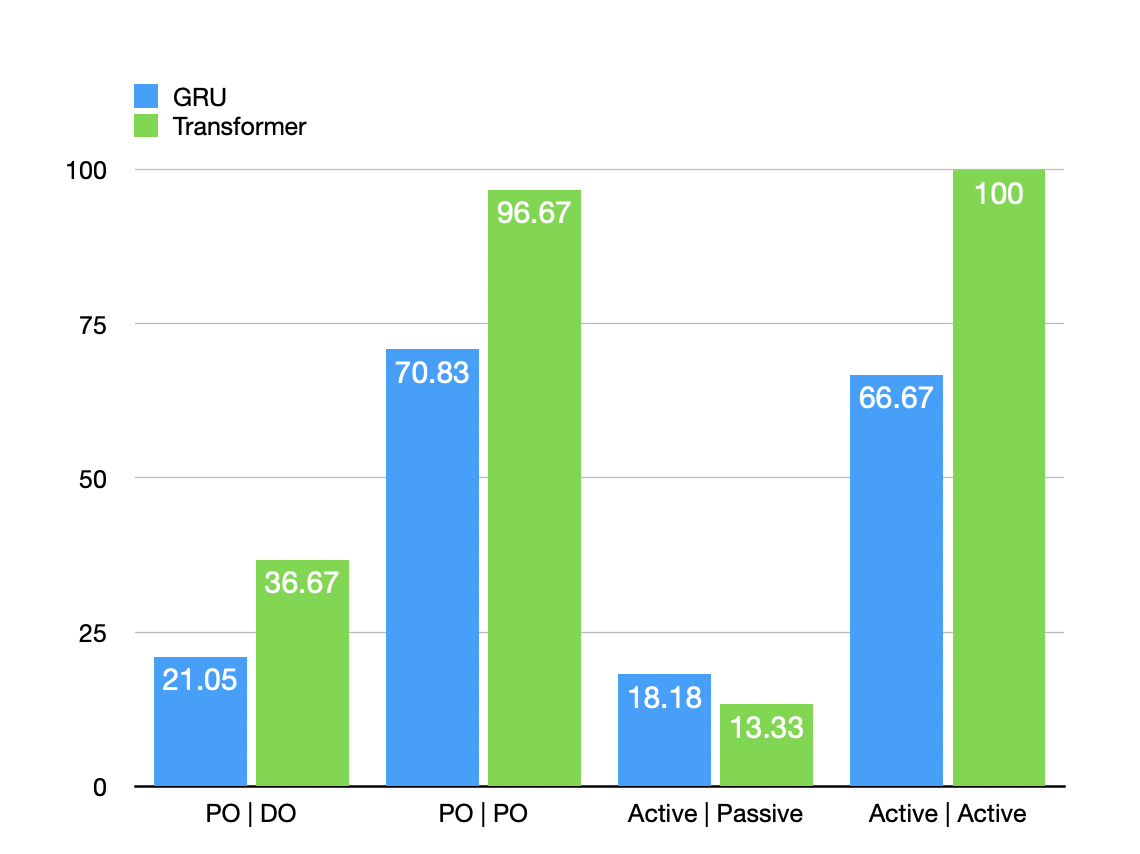}
\caption{
Priming Effect per Chunk: Proportion of correct cross-language priming chunks in the machine prediction results.}
\label{Priming_proportion}
\vspace*{-.15in}
\end{figure}

Since the sum of the normalized probabilities 
for the two target sentences is 1, we only need to consider the probability of one target type and compare 
it across
different priming types. The 
probability of another target type can be derived from this, i.e. $P_N(\overline{\text{Target}} | \text{Prime}) = 1 - P_N(\text{Target} | \text{Prime})$. By considering only one 
target type, we can directly compare the priming effects of the two priming types on
the specific target,
which is 
a key aspect of structural priming analysis.
The quantitative 
findings depicted in Figure~\ref{Priming_proportion} indicate that 
the transformer model generally outperforms GRU. Additionally,
a horizontal analysis 
of priming structural types
reveals that
machine predictions 
perform better with
active/passive structures compared to 
PO/DO structures.



\section{Summary and Conclusions}
This study evaluates cross-language structural priming effects in RNN and transformer models in 
a Chinese-English context. The models 
are trained 
on sentence pairs from both languages. 
Our research aims
to compare the structural priming abilities 
of different models. Even when using
the same training set,
which contains structurally primed sentences, RNNs and transformers still exhibit
differences in their ability
to achieve this effect. We find evidence for abstract crosslingual grammatical representations in 
these models, 
which operate
similarly to those found in 
prior research.

Our results show that BLEU scores
decrease as n-gram length
increases, 
consistent with
findings in sentence-similarity evaluation \cite{he_enhancing_2022}. Longer n-grams (e.g., bigrams and trigrams)
capture 
more specific 
contexts, making exact matches less likely unless the target sentence is very 
precise. Moreover, 
minor errors in word choice or sequence can disrupt the alignment of these 
n-grams.

Importantly, our results indicate that transformer models outperform RNNs in modeling Chinese-English structural priming, a finding that is intriguing given prior research. Traditionally, RNNs have been effective in modeling human sentence processing, 
explaining 
garden-path effects and structural priming through their sequential processing capabilities, which are thought to mirror aspects of human cognitive processing \cite{frank2021cross}. 

Our results show that the transformer model is more effective at preserving structural information than the RNN. The standardized accuracy rates for the transformer model exceed those of the RNN by 25.84\% for the PO structure and by 33.33\% for the active structure. This offers guidance for selecting base models in future computational linguistics research aimed at implementing or enhancing structural priming effects. This superiority of transformers raises questions about the efficacy of RNNs as human sentence processing models, especially if they are surpassed by a model considered less cognitively plausible. However, these results could also be seen as supporting the cognitive plausibility of transformers, particularly due to the attention mechanism.

While the concept of unlimited working memory in transformers 
seems implausible, some researchers argue that 
human working memory capacity is much smaller than traditionally estimated, limited to only two or three items. They suggest that language processing involves rapid, direct-access retrieval of items from memory \cite{lewis_computational_2006}, a process compatible with the attention mechanism in transformers. This mechanism assigns weights to 
past inputs based on their relevance to the current input, 
consistent with cue-based retrieval theories, 
where memory retrieval is influenced by the similarity of current cues to stored information \cite{parker2018cue}.

Our study on translation models extends the traditional RNN and Transformer comparisons in cognitive science, typically applied to language models for predictive coding. \citet{michaelov_structural_2023} have shown Transformers often better capture human language structure. While distinct from pure language modeling, our translation-focused approach offers insights into structural representations in neural networks and lays groundwork for refined language production models.

\section{Future Directions}
A promising future direction
is to develop
a model 
that generates
sentences based on new 
semantic concepts and thematic roles before and after priming. While 
challenging,
this approach could help mitigate
the lexical boost effect (see Limitations). 

Shifting
our focus from production to comprehension could also be fruitful. By measuring 
surprisal levels in models, we can 
explore
how structural priming influences 
comprehension, as suggested in recent studies \cite{merkx_human_2021}. 
Surprisal
quantifies the unexpectedness of a word in a given 
context, 
with lower values indicating higher
probability.
Consistently lower surprisal levels
at structurally complex points in sentences 
following priming. This
would suggest effective preparation by the priming process, offering
a way to explore 
the impact of
structural priming 
on language processing in model
without
the confounding effects of repeated vocabulary.

Additionally, 
evidence suggests an inverse relationship between the frequency of linguistic constructions and the magnitude of priming effects observed with those constructions \citep{jaeger_alignment_2013, kaschak_structural_2011}. For example, the double object (DO) construction is more common in American English than the prepositional object (PO) construction \cite{bock_persistence_2000}. Studies have shown that the less frequent PO construction exhibits stronger priming effects 
than the more frequent DO construction \cite{kaschak_structural_2011}. This aligns with theories of implicit learning in structural priming, where more frequently encountered structures are less ``surprising'' 
and thus generate weaker priming effects. 

To explore this further,
training models on corpora 
of American versus British English, which differ in their 
construction frequencies, could reveal whether
a similar inverse frequency effect is observed in computational models. This approach 
could shed light on
the dependency of structural priming on construction frequency, 
offering
deeper insights into how implicit learning processes are modeled computationally.

Additionally,
exploring
crowdsourcing as a method to enhance the sensitivity and grammaticality judgments of the test dataset could be valuable. By leveraging a diverse pool of contributors, 
this approach may provide
a wider range of evaluations and insights, 
potentially refining
our assessments and 
leading to more robust results.
 
\subsection*{Limitations}
\label{sec:limitation}

A limitation of the current study is that the Chinese-English priming effects observed in the models have not been directly compared with human data. Although existing evidence indicates a strong Chinese-English structural priming effect in both production and comprehension \citep{hsieh_structural_2017, Chen_2013}, equating the models’ ability to replicate cross-language priming with the structural ``correctness'' of their outputs may be somewhat simplistic. This 
underscores the need for future research that could involve
using the same stimuli with Mandarin-English bilinguals and making direct comparisons to human priming data. Such an approach would provide a more accurate assessment of the models’ alignment with human language processing.

Another
limitation 
is that our models 
cannot generate
sentences based on novel word concepts and thematic roles, such as the picture naming task in Figure~\ref{fig:figure_example}. Consequently, some critics may argue that what our models essentially do is translate from Chinese to English without generating new semantic content, as the semantic information remains consistent from the priming sentence to the output sentence. However, we maintain that the current study design validly assesses the priming effect, as 
the models must choose which sentence structure to use from among various structures that share the same semantic content---a choice influenced by the priming effect.

Nevertheless, 
we acknowledge that our design is susceptible to the ``lexical boost'' effect, where the structural priming effect is intensified when the same lexical head is repeated in both the prime and target sentences \cite{pickering_representation_1998}. For instance, if the target sentence is \textit{Alice gave Bob a book}, the priming effect is more pronounced if the prime sentence is 
\textit{Carl gave Danis a letter} 
rather than 
\textit{Alice showed Bob a book}.
Given that the semantic content remains constant across the prime and output sentences in our study, the observed priming effect 
may be artificially strengthened compared to what might be observed in a pure priming task.

Previous studies 
suggest that crosslingual structural priming might be affected by the asymmetry of training sources in certain language pairs \cite{michaelov_structural_2023}.  By measuring the probability shifts
for source and target sentences, we find such multilingual auto-regressive transformer 
models display evidence of abstract structural priming effects, although their performance
varies across different
scenarios.

\section*{Ethical Statement}

The current study adheres to the ethical standards set forth in the ACL Code of Ethics. The training dataset used in this research is open, publicly available, and does not include demographic or identity characteristics \citep{bright_xu_2019_3402023}.

Potential risks stem from
the fact that translations in the training data (a Chinese-English parallel sentence pair dataset) may not always be perfectly equivalent. Some words may carry cultural nuances that differ between Chinese and English. For example, the terms ''heshang'' and ''nígū'', translated as ``monk'' and ``nun,'' have specific cultural connotations in Chinese that differ from the perception of a ``monk'' in Western contexts, which is typically associated with Christian monasticism. These roles in Chinese Buddhism embody cultural and social aspects not fully captured by the Western terms, potentially leading to a loss of cultural meaning in translation.

Furthermore, while ChatGPT has been used to expand the test dataset, the authors have manually verified the output to ensure it remains unbiased. The potential risk of misuse of the computational model is low, as the encoders and decoders are designed to perform straightforward translation tasks and do not have the capability to self-generate harmful content.

\section*{Acknowledgments}
The last two authors are supported, in part, by DARPA Contract No. HR001121C0186. Any opinions, findings and conclusions or recommendations expressed in this material are those of the authors and do not necessarily reflect the views of the US Government.


\bibliography{custom,NLPClass.bib}

\end{document}